\documentclass[11pt,letterpaper]{article}
\usepackage{emnlp2017}
\usepackage{times}
\usepackage{latexsym}
\usepackage[pdftex]{graphicx}
\usepackage[tight,footnotesize]{subfigure}
\usepackage{amsmath}
\usepackage{amsfonts}

\DeclareMathOperator*{\argmin}{arg\,min}

\emnlpfinalcopy

\def\emnlppaperid{191}

\title{Unfolding and Shrinking Neural Machine Translation Ensembles}

\author{Felix Stahlberg \and Bill Byrne\\
{\tt	\{fs439,wjb31\}@cam.ac.uk} \\
         Department of Engineering\\
         University of Cambridge, UK}

\date{}

\begin{document}

\maketitle

\begin{abstract}
Ensembling is a well-known technique in neural machine translation (NMT) to improve system performance. Instead of a single neural net, multiple neural nets with the same topology are trained separately, and the decoder generates predictions by averaging over the individual models. Ensembling often improves the quality of the generated translations drastically. However, it is not suitable for production systems because it is cumbersome and slow. This work aims to reduce the runtime to be on par with a single system without compromising the translation quality. First, we show that the ensemble can be unfolded into a single large neural network which imitates the output of the ensemble system. We show that unfolding can already improve the runtime in practice since more work can be done on the GPU. We proceed by describing a set of techniques to shrink the unfolded network by reducing the dimensionality of layers. On Japanese-English we report that the resulting network has the size and decoding speed of a single NMT network but performs on the level of a 3-ensemble system.
\end{abstract}

\section{Introduction}

The top systems in recent machine translation evaluation campaigns on various language pairs use ensembles of a number of NMT systems~\cite{sys-wmt16,sys-uedin-wmt16,nmt-char-noseg,sys-neubig-wat16,sys-google,sys-kyoto-wat16,sys-qcri}. Ensembling~\cite{ml-ensembles,nn-ensembles} of neural networks is a simple yet very effective technique to improve the accuracy of NMT. The decoder makes use of $K$ NMT networks which are either trained independently~\cite{nmt-sutskever,nmt-char-noseg,sys-neubig-wat16,sys-google} or share some amount of training iterations~\cite{nmt-bpe,sys-uedin-wmt16,sys-kyoto-wat16,sys-qcri}. The ensemble decoder computes predictions from each of the individual models which are then combined using the arithmetic average~\cite{nmt-sutskever} or the geometric average~\cite{sys-kyoto-wat16}.

\begin{figure*}[!t]
\centering
\subfigure[Single network 1.]{\label{fig:network1}\includegraphics[width=0.32\linewidth]{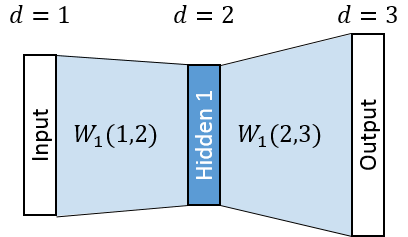}}
\subfigure[Single network 2.]{\label{fig:network2}\includegraphics[width=0.32\linewidth]{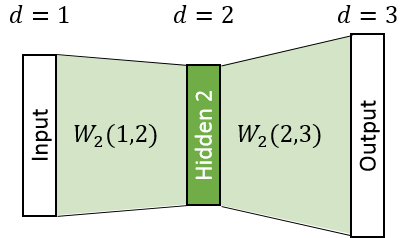}}
\subfigure[Unfolded network.]{\label{fig:network-combi}\includegraphics[width=0.32\linewidth]{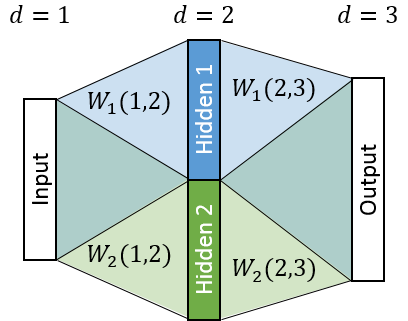}}

\caption{Unfolding mimics the output of the ensemble of two single layer feedforward networks.}
\label{fig:unfolding}
\end{figure*}

Ensembling consistently outperforms single NMT by a large margin. However, the decoding speed is significantly worse since the decoder needs to apply $K$ NMT models rather than only one. Therefore, a recent line of research transfers the idea of {\em knowledge distillation}~\cite{kd-original,kd-hinton} to NMT and trains a smaller network (the student) by minimizing the cross-entropy to the output of the ensemble system (the teacher)~\cite{kd-nmt,kd-nmt-ensemble}. This paper presents an alternative to knowledge distillation as we aim to speed up decoding to be comparable to single NMT while retaining the boost in translation accuracy from the ensemble. In a first step, we describe how to construct a single large neural network which imitates the output of an ensemble of multiple networks with the same topology. We will refer to this process as {\em unfolding}. GPU-based decoding with the unfolded network is often much faster than ensemble decoding since more work can be done on the GPU. In a second step, we explore methods to reduce the size of the unfolded network. This idea is justified by the fact that ensembled neural networks are often over-parameterized and have a large degree of redundancy~\cite{sparsify-obd,sparsify-obs,sparsify-datafree}. Shrinking the unfolded network leads to a smaller model which consumes less space on the disk and in the memory; a crucial factor on mobile devices. More importantly, the decoding speed on all platforms benefits greatly from the reduced number of neurons.
We find that the dimensionality of linear embedding layers in the NMT network can be reduced heavily by low-rank matrix approximation based on singular value decomposition (SVD). This suggest that high dimensional embedding layers may be needed for training, but do not play an important role for decoding. The NMT network, however, also consists of complex layers like gated recurrent units~\cite[GRUs]{nmt-gru} and attention~\cite{nmt-bahdanau}. Therefore, we introduce a novel algorithm based on linear combinations of neurons which can be applied either during training ({\em data-bound}) or directly on the weight matrices without using training data ({\em data-free}). We report that with a mix of the presented shrinking methods we are able to reduce the size of the unfolded network to the size of the single NMT network while keeping the boost in BLEU score from the ensemble. Depending on the aggressiveness of shrinking, we report either a gain of 2.2 BLEU at the same decoding speed, or a 3.4$\times$ CPU decoding speed up with only a minor drop in BLEU compared to the original single NMT system. Furthermore, it is often much easier to stage a single NMT system than an ensemble in a commercial MT workflow, and it is crucial to be able to optimize quality at specific speed and memory constraints. Unfolding and shrinking address these problems directly.

\section{Unfolding $K$ Networks into a Single Large Neural Network}
\label{sec:unfolding}

The first concept of our approach is called {\em unfolding}. Unfolding is an alternative to ensembling of multiple neural networks with the same topology. Rather than averaging their predictions, unfolding constructs a single large neural net out of the individual models which has the same number of input and output neurons but larger inner layers. Our main motivation for unfolding is to obtain a single network with ensemble level performance which can be shrunk with the techniques in Sec.~\ref{sec:shrinking}. 

Suppose we ensemble two single layer feedforward neural nets as shown in Fig.~\ref{fig:unfolding}. Normally, ensembling is implemented by performing an isolated forward pass through the first network (Fig.~\ref{fig:network1}), another isolated forward pass through the second network (Fig.~\ref{fig:network2}), and averaging the activities in the output layers of both networks. This can be simulated by merging both networks into a single large network as shown in Fig.~\ref{fig:network-combi}. The first neurons in the hidden layer of the combined network correspond to the hidden layer in the first single network, and the others to the hidden layer of the second network. A single pass through the combined network yields the same output as the ensemble if the output layer is linear (up to a factor 2). The weight matrices in the unfolded network can be constructed by stacking the corresponding weight matrices (either horizontally or vertically) in network 1 and 2. This kind of aggregation of multiple networks with the same topology is not only possible for single-layer feedforward architectures but also for complex networks consisting of multiple GRU layers and attention.

\begin{figure*}[!t]
\centering
\small

\begin{equation*}
W'(d_1,d_2) =  \begin{cases}

    \begin{pmatrix}
  W_1(d_1,d_2) & 0 & \cdots & 0 \\
  0 & W_2(d_1,d_2)  & \vdots & \vdots \\
  \vdots  & \cdots  & \ddots & 0 \\
  0  & \cdots  &  & W_K(d_1,d_2) \\
 \end{pmatrix}       & \quad \text{if } d_1\in \text{InnerLayers} \text{ and } d_2\in \text{InnerLayers}\\

    \frac{1}{K}\begin{pmatrix}
  W_1(d_1,d_2) \\
  \vdots \\
  W_K(d_1,d_2) \\
 \end{pmatrix}       & \quad \text{if } d_1\in\text{InnerLayers} \text{ and } d_2\notin\text{InnerLayers}\\

    \begin{pmatrix}
  W_1(d_1,d_2) & \cdots & W_K(d_1,d_2) \\
 \end{pmatrix} & \quad \text{if } d_1\notin\text{InnerLayers} \text{ and } d_2\in\text{InnerLayers}\\
 
  \end{cases}
\end{equation*}
\caption{General formula for unfolding weight matrices. The set $\text{InnerLayers}:=[2,D-1]$ includes all layers except the input, output, and bias layer.}
\label{fig:unfolding-eq}
\end{figure*}


For a formal description of unfolding we address layers with indices $d=0,1,\dots,D$. The special layer $0$ has a single neuron for modelling bias vectors. Layer $1$ holds the input neurons and layer $D$ is the output layer. We denote the size of a layer in the individual models as $s(d)$. When combining $K$ networks, the layer size $s'(d)$ in the unfolded network is increased by factor $K$ if $d$ is an inner layer, and equal to $s(d)$ if $d$ is the input or output layer.
We denote the weight matrix between two layers $d_1,d_2\in [0,D]$ in the $k$-th individual model ($k\in [1,K]$) as $W_k(d_1,d_2)\in \mathbb{R}^{s(d_1)\times s(d_2)}$, and the corresponding weight matrix in the unfolded network as $W'(d_1,d_2)\in \mathbb{R}^{s'(d_1)\times s'(d_2)}$. We explicitly allow $d_1$ and $d_2$ to be non-consecutive or reversed to be able to model recurrent networks. We use the zero-matrix if layers $d_1$ and $d_2$ are not connected. The construction of the unfolded weight matrix $W'(d_1,d_2)$ from the individual matrices $W_k(d_1,d_2)$ depends on whether the connected layers are inner layers or not. The complete formula is listed in Fig.~\ref{fig:unfolding-eq}.

Unfolded NMT networks approximate but do not exactly match the output of the ensemble due to two reasons. First, the unfolded network synchronizes the attentions of the individual models. Each decoding step in the unfolded network computes a single attention weight vector. In contrast, ensemble decoding would compute one attention weight vector  for each of the $K$ input models. A second difference is that the ensemble decoder first applies the softmax at the output layer, and then averages the prediction probabilities. The unfolded network averages the neuron activities (i.e.\ the logits) first, and then applies the softmax function. Interestingly, as shown in Sec.~\ref{sec:results}, these differences do not have any impact on the BLEU score but yield potential speed advantages of unfolding since the computationally expensive softmax layer is only applied once.

\section{Shrinking the Unfolded Network}
\label{sec:shrinking}

After constructing the weight matrices of the unfolded network we reduce the size of it by iteratively shrinking layer sizes. In this section we denote the incoming weight matrix of the layer to shrink as $U\in \mathbb{R}^{m_{in}\times m}$ and the outgoing weight matrix as $V\in \mathbb{R}^{m\times m_{out}}$. Our procedure is inspired by the method of Srinivas and Babu~\shortcite{sparsify-datafree}. They propose a criterion for removing neurons in inner layers of the network based on two intuitions. First, similarly to Hebb's learning rule, they detect redundancy by the principle {\em neurons which fire together, wire together}. 
If the incoming weight vectors $U_{:,i}$ and $U_{:,j}$ are exactly the same for two neurons $i$ and $j$, we can remove the neuron $j$ and add its outgoing connections to neuron $i$ ($V_{i,:}\gets V_{i,:}+V_{j,:}$) without changing the output.\footnote{We denote the $i$-th row vector of a matrix $A$ with $A_{i,:}$ and the $i$-th column vector as $A_{:,i}$.} This holds since the activity in neuron $j$ will always be equal to the activity in neuron $i$. In practice, Srinivas and Babu use a distance measure based on the difference of the incoming weight vectors to search for similar neurons as exact matches are very rare.

The second intuition of the criterion used by Srinivas and Babu~\shortcite{sparsify-datafree} is that neurons with small outgoing weights contribute very little overall. Therefore, they search for a pair of neurons $i,j\in [1,m]$ according the following term and remove the $j$-th neuron.\footnote{Note that the criterion in Eq.~\ref{eq:babu} generalizes the criterion of Srinivas and Babu~\shortcite{sparsify-datafree} to multiple outgoing weights. Also note that Srinivas and Babu~\shortcite{sparsify-datafree} propose some heuristic improvements to this criterion. However, these heuristics did not work well in our NMT experiments.}

\begin{equation}
\label{eq:babu}
\argmin_{i,j\in [1,m]} ||U_{:,i}-U_{:,j}||_2^2 ||V_{j,:}||_2^2 
\end{equation}

Neuron $j$ is selected for removal if (1) there is another neuron $i$ which has a very similar set of incoming weights and if (2) $j$ has a small outgoing weight vector. Their criterion is {\em data-free} since it does not require any training data. For further details we refer to Srinivas and Babu~\shortcite{sparsify-datafree}.

\subsection{Data-Free Neuron Removal}
\label{sec:data-free}

Srinivas and Babu~\shortcite{sparsify-datafree} propose to add the outgoing weights of $j$ to the weights of a similar neuron $i$ to compensate for the removal of $j$. However, we have found that this approach does not work well on NMT networks. We propose instead to compensate for the removal of a neuron by a linear combination of the remaining neurons in the layer. Data-free shrinking assumes for the sake of deriving the update rule that the neuron activation function is linear. We now ask the following question: How can we compensate as well as possible for the loss of neuron $j$ such that the impact on the output of the  whole network is minimized? Data-free shrinking represents the incoming weight vector of neuron $j$ ($U_{:,j}$) as linear combination of the incoming weight vectors of the other neurons. The linear factors can be found by satisfying the following linear system:
\begin{equation}
\label{eq:approx}
U_{:,\lnot j} \lambda = U_{:,j}
\end{equation}
where $U_{:,\lnot j}$ is matrix $U$ without the $j$-th column. In practice, we use the method of ordinary least squares  to find $\lambda$ because the system may be overdetermined. The idea is that if we mix the outputs of all neurons in the layer by the $\lambda$-weights, we get the output of the $j$-th neuron. The row vector $V_{j,:}$ contains the contributions of the $j$-th neuron to each of the neurons in the next layer. Rather than using these connections, we approximate their effect by adding some weight to the outgoing connections of the other neurons. How much weight depends on $\lambda$ and the outgoing weights $V_{j,:}$. The factor $D_{k,l}$ which we need to add to the outgoing connection of the $k$-th neuron to compensate for the loss of the $j$-th neuron on the $l$-th neuron in the next layer is:

\begin{equation}
\label{eq:update-delta}
D_{k,l}=\lambda_k V_{j,l}
\end{equation}

Therefore, the update rule for $V$ is:

\begin{equation}
\label{eq:update}
V \gets V + D
\end{equation}

In the remainder we will refer to this method as {\em data-free} shrinking. Note that we recover the update rule of Srinivas and Babu~\shortcite{sparsify-datafree} 
by setting $\lambda$ to the $i$-th unit vector. Also note that the error introduced by our shrinking method is due to the fact that we ignore the non-linearity, and that the solution for $\lambda$ may not be exact. The method is error-free on linear layers as long as the residuals of the least-squares analysis in Eq.~\ref{eq:approx} are zero.

\paragraph{GRU layers}

The terminology of {\em neurons} needs some further elaboration for GRU layers which rather consist of update and reset gates and states~\cite{nmt-gru}. On GRU layers, we treat the states as neurons, i.e.\ the $j$-th neuron refers to the $j$-th entry in the GRU state vector. Input connections to the gates are included in the incoming weight matrix $U$ for estimating $\lambda$ in Eq.~\ref{eq:approx}.  Removing neuron $j$ in a GRU layer means deleting the $j$-th entry in the states and both gate vectors.

\subsection{Data-Bound Neuron Removal}
\label{sec:data-bound}

Although we find our data-free approach to be a substantial improvement over the methods of Srinivas and Babu~\shortcite{sparsify-datafree} on NMT networks, it still leads to a non-negligible decline in BLEU score when applied to recurrent GRU layers.  Our data-free method uses the incoming weights to identify similar neurons, i.e.\ neurons expected to have similar activities.  This works well enough for simple layers, but the interdependencies between the states and the gates inside gated layers like GRUs or LSTMs are  complex enough that redundancies cannot be found simply by looking for similar weights. In the spirit of Babaeizadeh et al.\ \shortcite{sparsify-noiseout}, our {\em data-bound} version records neuron activities during training to estimate $\lambda$. We compensate for the removal of the $j$-th neuron by using a linear combination of the output of remaining neurons with similar activity patterns. In each layer, we prune $40$ neurons each $450$ training iterations until the target layer size is reached. Let $A$ be the matrix which holds the records of neuron activities in the layer since the last removal. For example, for the decoder GRU layer, a batch size of 80, and target sentence lengths of 20, $A$ has $20\cdot 80 \cdot 450=720K$ rows and $m$ (the number of neurons in the layer) columns.\footnote{In practice, we use a random sample of 50K rows rather than the full matrix to keep the complexity of the least-squares analysis under control.} Similarly to Eq.~\ref{eq:approx} we find interpolation weights $\lambda$ using the method of least squares on the following linear system.

\begin{equation}
\label{eq:approx-databound}
A_{:,\lnot j} \lambda = A_{:,j}
\end{equation}

The update rule for the outgoing weight matrix is the same as for our data-free method (Eq.~\ref{eq:update}). The key difference between data-free and data-bound shrinking is the way $\lambda$ is estimated. Data-free shrinking uses the similarities between incoming weights, and data-bound shrinking uses neuron activities recorded during training. Once we select a neuron to remove, we estimate $\lambda$, compensate for the removal, and proceed with the shrunk network. Both methods are prior to any decoding and result in shrunk parameter files which are then loaded to the decoder. Both methods remove neurons rather than single weights.

The data-bound algorithm runs gradient-based optimization on the unfolded network. We use the AdaGrad~\cite{train-adagrad} step rule, a small learning rate of 0.0001, and aggressive step clipping at 0.05 to avoid destroying useful weights which were learned in the individual networks prior to the construction of the unfolded network.

Our data-bound algorithm uses a data-bound version of the neuron selection criterion in Eq.~\ref{eq:babu} which operates on the activity matrix $A$. We search for the pair $i,j\in [1,m]$ according the following term and remove neuron $j$.

\begin{equation}
\label{eq:criterion-databound}
\argmin_{i,j\in [1,m]}||A_{:,i}-A_{:,j}||_2^2 ||A_{:,j}||_2^2
\end{equation}

\subsection{Shrinking Embedding Layers with SVD}
\label{sec:svd}

The standard attention-based NMT network architecture~\cite{nmt-bahdanau} includes three linear layers: the embedding layer in the encoder, and the output and feedback embedding layers in the decoder. We have found that linear layers are particularly easy to shrink using low-rank matrix approximation. As before we denote the incoming weight matrix as $U\in \mathbb{R}^{m_{in}\times m}$ and the outgoing weight matrix as $V\in \mathbb{R}^{m\times m_{out}}$. Since the layer is linear, we could directly connect the previous layer with the next layer using the product of both weight matrices $X=U\cdot V$. However, $X$ may be very large. Therefore, we approximate $X$ as a product of two low rank matrices $Y\in\mathbb{R}^{m_{in}\times m'}$ and $Z\in\mathbb{R}^{m' \times m_{out}}$ ($X\approx YZ$) where $m'\ll m$ is the desired layer size. A very common way to find such a matrix factorization is using truncated singular value decomposition (SVD). The layer is eventually shrunk by replacing $U$ with $Y$ and $V$ with $Z$.

\begin{table*}
\small
\centering
\begin{tabular}{@{\hspace{0em}}r@{\hspace{0.1em}}|l||cc|c|ccc||r|rr|}
\cline{3-8}
\multicolumn{2}{c||}{} & \multicolumn{6}{c||}{\bf Shrinking Methods} & \multicolumn{3}{|c}{} \\
\cline{2-11}
& \multicolumn{1}{|c||}{\bf Base} & \multicolumn{2}{|c|}{\bf Encoder} & \bf Attention & \multicolumn{3}{|c||}{\bf Decoder} & \multicolumn{1}{|c|}{\bf Size} & \multicolumn{2}{|c|}{\bf BLEU} \\
& & \bf Embed. & \bf GRUs & \bf Match & \bf GRU & \bf Maxout & \bf Embeds. &  \bf Factor & \bf dev & \bf test \\ 
\cline{2-11}
\footnotesize{(a)} & Single & - & - & - & - & - & - & 1.00 & 20.8 & 23.5 \\
\footnotesize{(b)} & 2-Ens. & - & - & - & - & - & - & 2$\times$1.00 & 22.7 & 25.2 \\
\cline{2-11}
\footnotesize{(c)} & 2-Unfold & SVD & - & - & - & - & SVD & 1.85 & 22.7 & 25.1 \\
\footnotesize{(d)} & 2-Unfold & SVD & - & Data-Free & - & - & SVD & 1.77 & 22.7 & 25.1 \\
\footnotesize{(e)} & 2-Unfold & SVD & Data-Free & Data-Free & Data-Free & - & SVD & 1.05 & 21.6 & 24.2 \\ 
\footnotesize{(f)} & 2-Unfold & SVD & Data-Bound & Data-Free & Data-Bound & - & SVD & 1.05 & 22.4 & \bf 25.3 \\
\cline{2-11}
\footnotesize{(g)} & 2-Unfold & SVD & Data-Bound & Data-Free & Data-Bound & Data-Free & SVD & 1.00 & 16.9 & 19.3 \\
\footnotesize{(h)} & 2-Unfold & SVD & Data-Bound & Data-Free & Data-Bound & Data-Bound & SVD & 1.00 & 21.9 & 24.6 \\
\cline{2-11}
\end{tabular}
\caption{\label{tab:layer-wise} Shrinking layers of the unfolded network on Ja-En to their original size.}
\end{table*}

\section{Results}
\label{sec:results}

The individual NMT systems we use as source for constructing the unfolded networks are trained using AdaDelta~\cite{train-adadelta} on the Blocks/Theano implementation~\cite{train-blocks,train-theano} of the standard attention-based NMT model~\cite{nmt-bahdanau} with: 1000 dimensional GRU layers~\cite{nmt-gru} in both the decoder and bidrectional encoder; a single maxout output layer~\cite{maxout}; and 620 dimensional embedding layers. We follow Sennrich et al.~\shortcite{nmt-bpe} and use subword units based on byte pair encoding rather than words as modelling units. Our SGNMT decoder~\cite{sgnmt}\footnote{`vanilla' decoding strategy} 
with a beam size of 12 is used in all experiments. Our primary corpus is the Japanese-English (Ja-En) ASPEC data set~\cite{data-aspec}. We select a subset of 500K sentence pairs to train our models as suggested by Neubig et al.~\shortcite{sys-neubig-wat15}. We report cased BLEU scores calculated with Moses' \texttt{multi-bleu.pl} to be strictly comparable to the evaluation done in the Workshop of Asian Translation (WAT). We also apply our method to the WMT data set for English-German (En-De), using the {\em news-test2014} as a development set, and keeping {\em news-test2015} and {\em news-test2016} as test sets. En-De BLEU scores are computed using \texttt{mteval-v13a.pl} as in the WMT evaluation. We set the vocabulary sizes to 30K for Ja-En and 50K for En-De. We also report the {\em size factor} for each model which is the total number of model parameters (sum of all weight matrix sizes) divided by the number of parameters in the original NMT network (86M for Ja-En and 120M for En-De). 
We choose a widely used, simple ensembling method (prediction averaging) as our baseline. We feel that the prevalence of this method makes it a reasonable baseline for our experiments.

\paragraph{Shrinking the Unfolded Network}

First, we investigate which shrinking methods are effective for which layers. Tab.~\ref{tab:layer-wise} summarizes our results on a 2-unfold network for Ja-En, i.e.\ two separate NMT networks are combined in a single large network as described in Sec.~\ref{sec:unfolding}. The layers in the combined network are shrunk to the size of the original networks using the methods discussed in Sec.~\ref{sec:shrinking}. 

Shrinking the linear embedding layers with SVD (Sec.~\ref{sec:svd}) is very effective. The unfolded model with shrunk embedding layers performs at the same level as the ensemble (compare rows (b) and (c)). In our initial experiments, we applied the method of Srinivas and Babu~\shortcite{sparsify-datafree} to shrink the other layers, but their approach performed very poorly on this kind of network: the BLEU score dropped down to 15.5 on the development set when shrinking all layers except the decoder maxout and embedding layers, and to 9.9 BLEU when applying their method only to embedding layers.\footnote{Results with the original method of Srinivas and Babu~\shortcite{sparsify-datafree} are not included in Tab.~\ref{tab:layer-wise}.} Row (e) in Tab.~\ref{tab:layer-wise} shows that our data-free algorithm from Sec.~\ref{sec:data-free} is better suited for shrinking the GRU and attention layers, leading to a drop of only 1 BLEU point compared to the ensemble (b) (i.e.\ 0.8 BLEU better than the single system (a)). However, using the data-bound version of our shrinking algorithm (Sec.~\ref{sec:data-bound}) for the GRU layers performs best.\footnote{If we apply different methods to different layers of the same network, we first apply SVD-based shrinking, then the data-free method, and finally the data-bound method.} The shrunk model yields about the same BLEU score as the ensemble on the test set (25.2 in (b) and 25.3 in (f)). Shrinking the maxout layer remains more of a challenge (rows (g) and (h)), but the number of parameters in this layer is small. Therefore, shrinking all layers except the maxout layer leads to almost the same number of parameters (factor 1.05 in row (f)) as the original NMT network (a),  and thus to a similar storage size, memory consumption, and decoding speed, but with a 1.8 BLEU gain.
Based on these results we fix the shrinking method used for each layer for all remaining experiments as follows: We shrink linear embedding layers with our SVD-based method, GRU layers with our data-bound method, the attention layer with our data-free method, and do not shrink the maxout layer.

\begin{table}
\small
\centering
\begin{tabular}{@{\hspace{0em}}r@{\hspace{0.1em}}|cc|rr|}
\cline{2-5} 
& \multicolumn{2}{|c|}{\bf Compensation Method} & \multicolumn{2}{c|}{\bf BLEU} \\
& \bf Linear Combination & \bf SGD & \bf dev & \bf test \\
\cline{2-5} 
\footnotesize{(a)} & &  & 16.3 & 18.0 \\
\footnotesize{(b)} & $\checkmark$ & & 22.1 & 24.3 \\
\footnotesize{(c)} & & $\checkmark$ & 21.7 & 24.4 \\
\footnotesize{(d)} & $\checkmark$ & $\checkmark$ & 22.4 & 25.3\\
\cline{2-5}
\end{tabular}
\caption{\label{tab:data-bound} Compensating for neuron removal in the data-bound algorithm. Row (d) corresponds to row (f) in Tab.~\ref{tab:layer-wise}.}
\end{table}

\begin{table*}
\small
\centering
\begin{tabular}{@{\hspace{0em}}r@{\hspace{0.1em}}|l|rr|r|rr|}
\cline{2-7}
& \multicolumn{1}{|c|}{\bf System} & \multicolumn{2}{|c|}{\bf Words/Min.}  & \multicolumn{1}{|c|}{\bf Size} & \multicolumn{2}{|c|}{\bf BLEU} \\
&  & \multicolumn{1}{c}{\bf CPU} & \multicolumn{1}{c|}{\bf GPU} & \multicolumn{1}{|c|}{\bf Factor} & \multicolumn{1}{|c}{\bf dev} & \multicolumn{1}{c|}{\bf test} \\
\cline{2-7}
\footnotesize{(a)} & Single & 323.4  & 2993.6 & 1.00 & 20.8 & 23.5 \\
\cline{2-7}
\footnotesize{(b)} & 2-Ensemble & 163.7 & 1641.1 &  2 $\times$ 1.00 & 22.7 & 25.2 \\
\footnotesize{(c)} & 2-Unfold, shrunk embed.\& attention & 157.2 & 2592.2  & 1.77 & 22.7 & 25.1 \\
\footnotesize{(d)} & 2-Unfold, shrunk all except maxout & 308.3 & 2961.4 & 1.05 & 22.4 &  25.3 \\
\cline{2-7}
\footnotesize{(e)} & 3-Ensemble & 110.9 & 1158.2 &  3 $\times$ 1.00 & 23.4 & 25.9 \\
\footnotesize{(f)} & 3-Unfold, shrunk embed.\& attention & 95.4 & 2182.1  & 2.99 & 23.2 & 25.9 \\
\footnotesize{(g)} & 3-Unfold, shrunk all except maxout & 301.6 & 3024.4 & 1.09 & 22.2 & 25.3 \\
\cline{2-7}
\end{tabular}
\caption{\label{tab:time} Time measurements on Ja-En. Layers are shrunk to their size in the original NMT model.}
\end{table*}

Our data-bound algorithm from Sec.~\ref{sec:data-bound} has two mechanisms to compensate for the removal of a neuron. First, we use a linear combination of the remaining neurons to update the outgoing weight matrix by imitating its activations (Eq.~\ref{eq:update}). Second, stochastic gradient descent (SGD) fine-tunes all weights during this process. Tab.~\ref{tab:data-bound} demonstrates that both mechanisms are crucial for minimizing the effect of shrinking on the BLEU score.

\paragraph{Decoding Speed}

Our testing environment is an Ubuntu 16.04 with Linux 4.4.0 kernel, 32 GB RAM, an Intel\textsuperscript{\textregistered} Core i7-6700 CPU at 3.40 GHz
and an Nvidia GeForce GTX Titan X GPU. CPU decoding uses a single thread. We used the first 500 sentences of the Ja-En WAT development set for the time measurements.

Our results in Tab.~\ref{tab:time} show that decoding with ensembles (rows (b) and (e)) is slow: combining the predictions of the individual models on the CPU is computationally expensive, and ensemble decoding requires $K$ passes through the softmax layer which is also computationally expensive. Unfolding the ensemble into a single network and shrinking the embedding and attention layers improves the runtimes on the GPU significantly without noticeable impact on BLEU (rows (c) and (f)). This can be attributed to the fact that unfolding can reduce the communication overhead between CPU and GPU. Comparing rows (d) and (g) with row (a) reveals that shrinking the unfolded networks even further speeds up CPU and GPU decoding almost to the level of single system decoding. However, more aggressive shrinking yields a BLEU score of 25.3 when combining three systems (row (g)) -- 1.8 BLEU better than the single system, but 0.6 BLEU worse than the 3-ensemble. Therefore, we will investigate the impact of shrinking on the different layers in the next sections more thoroughly.

\paragraph{Degrees of Redundancy in Different Layers}

\begin{figure}[!t]
\centering
\includegraphics[width=1\linewidth]{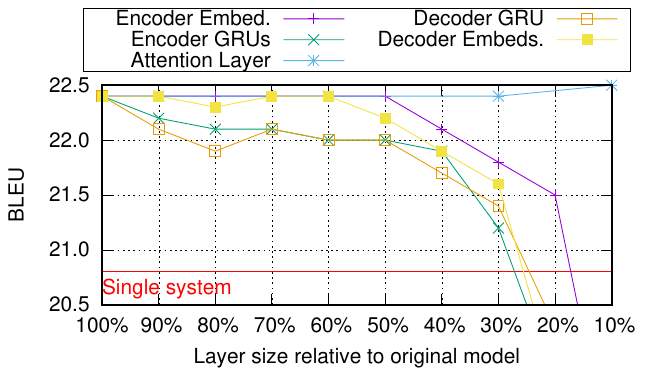}
\caption{Impact of shrinking on the BLEU score.}
\label{fig:bleu-over-shrinking}
\end{figure}

We applied our shrinking methods to isolated layers in the 2-Unfold network of Tab.~\ref{tab:layer-wise} (f). Fig.~\ref{fig:bleu-over-shrinking} plots the BLEU score when isolated layers are shrunk even below their size in the original NMT network. The attention layer is very robust against shrinking and can be reduced to 100 neurons (10\% of the original size) without impacting the BLEU score. The embedding layers can be reduced to 60\% but are sensitive to more aggressive pruning. Shrinking the GRU layers affects the BLEU score the most but still outperforms the single system when the GRU layers are shrunk to 30\%.

\paragraph{Adjusting the Target Sizes of Layers}

\begin{table}
\small
\centering
\begin{tabular}{|l|r|rrr|}
\hline 
 & \bf Single & \multicolumn{3}{c|}{\bf 3-Unfold} \\
 &  & \multicolumn{1}{c}{\bf Normal} & \multicolumn{1}{c}{\bf Small}  & \multicolumn{1}{c|}{\bf Tiny} \\
\hline
Enc.\ Embed. & 620 & 410 & 310 & 170 \\
Enc.\ GRUs & 1000 & 1300 & 580 & 580 \\
Attention & 1000 & 100 & 100 & 100 \\
Dec.\ GRU & 1000 & 1350 & 590 & 590 \\
Dec.\ Maxout & 500 & 1500 & 1500 & 1500 \\
Dec.\ Embeds. & 620 & 430 & 320 & 170 \\
\hline
Size Factor & 1.00 & 1.00 & 0.50 & 0.33 \\
\hline
\end{tabular}
\caption{\label{tab:setups} Layer sizes of our setups for Ja-En.}
\end{table}

\begin{table}
\small
\centering
\begin{tabular}{@{\hspace{0em}}r@{\hspace{0.1em}}|l|rr|rr|}
\cline{2-6} 
& \multicolumn{1}{|c|}{\bf System} & \multicolumn{2}{|c|}{\bf Words/Min.}  &  \multicolumn{2}{|c|}{\bf BLEU} \\
& & \multicolumn{1}{|c}{\bf CPU} & \multicolumn{1}{c|}{\bf GPU} &  \multicolumn{1}{|c}{\bf dev} & \multicolumn{1}{c|}{\bf test} \\
\cline{2-6} 
\footnotesize{(a)} & Single & 323.4 & 2993.6  & 20.8 & 23.5 \\
\footnotesize{(b)} & 3-Ensemble & 110.9 & 1158.2  & 23.4 & 25.9 \\
\cline{2-6} 
\footnotesize{(c)} & 3-Unfold-Normal & 445.2 & 3071.1 & 22.9 & 25.7 \\
\footnotesize{(d)} & 3-Unfold-Small & 946.1 & 3572.0 & 21.7 & 23.9 \\
\footnotesize{(e)} & 3-Unfold-Tiny & 1102.5 & 3483.7 &  20.6 & 23.2 \\
\cline{2-6} 
\end{tabular}
\caption{\label{tab:time-setups} Our best models on Ja-En.}
\end{table}

Based on our previous experiments we revise our approach to shrink the 3-Unfold system in Tab.~\ref{tab:time}. Instead of shrinking all layers except the maxout layer to the same degree, we adjust the aggressiveness of shrinking for each layer. We suggest three different setups ({\em Normal}, {\em Small}, and {\em Tiny}) with the layer sizes specified in Tab.~\ref{tab:setups}. {\em 3-Unfold-Normal} has the same number of parameters as the original NMT networks (size factor: 1.0), {\em 3-Unfold-Small} is only half their size (size factor: 0.5), and {\em 3-Unfold-Tiny} reduces the size by two thirds (size factor: 0.33). When comparing rows (a) and (c) in Tab.~\ref{tab:time-setups} we observe that  {\em 3-Unfold-Normal} yields a gain of 2.2 BLEU with respect to the original single system and a slight improvement in decoding speed at the same time.\footnote{To validate that the gains come from ensembling and unfolding and not from the layer sizes in {\em 3-Unfold-Normal} we trained a network from scratch with the same dimensions. This network performed similarly to our {\em Single} system.} Networks with the size factor 1.0 like {\em 3-Unfold-Normal} are very likely to yield about the same decoding speed as the {\em Single} network regardless of the decoder implementation, machine learning framework, and hardware. Therefore, we think that similar results are possible on other platforms as well.

CPU decoding speed directly benefits even more from smaller setups -- {\em 3-Unfold-Tiny} is only 0.3 BLEU worse than {\em Single} but decoding on a single CPU is 3.4 times faster (row (a) vs.\ row (e) in Tab.~\ref{tab:time-setups}). This is of great practical use: batch decoding with only two CPU threads surpasses production speed which is often set to 2000 words per minute~\cite{smt-speed}. Our initial experiments in Tab.~\ref{tab:time-setups-ende} suggest that the {\em Normal} setup is applicable to En-De as well, with substantial improvements in BLEU compared to {\em Single} with about the same decoding speed.

\begin{table}
\small
\centering
\begin{tabular}{|l|r|rrr|}
\hline 
\multicolumn{1}{|c|}{\bf System} & \multicolumn{1}{|c|}{\bf Wrds/Min.}  & \multicolumn{3}{|c|}{\bf BLEU on news-test*} \\
 & \multicolumn{1}{|c|}{\bf (GPU)} &  \multicolumn{1}{|c}{\bf 2014} & \multicolumn{1}{c}{\bf 2015} &  \multicolumn{1}{c|}{\bf 2016} \\
\hline
Single & 2128.7 & 19.6 & 21.9 & 24.6 \\
2-Ensemble & 1135.3 & 20.5 & 22.9 & 26.1 \\
\hline
2-Unfold-Norm. & 2099.1  & 20.7 & 23.1 & 25.8 \\
\hline
\end{tabular}
\caption{\label{tab:time-setups-ende} Our best models on En-De.}
\end{table}

\section{Related Work}

The idea of pruning neural networks to improve the compactness of the models dates back more than 25 years~\cite{sparsify-obd}. The literature is therefore vast~\cite{sparsify-review}. One line of research aims to remove unimportant network connections. The connections can be selected for deletion based on the second-derivative of the training error with respect to the weight~\cite{sparsify-obd,sparsify-obs}, or by a threshold criterion on its magnitude~\cite{sparsify-threshold}. See et al.~\shortcite{sparsify-nmt} confirmed a high degree of weight redundancy in NMT networks.

In this work we are interested in removing neurons rather than single connections since we strive to shrink the unfolded network such that it resembles the layout of an individual model. We argued in Sec.~\ref{sec:results} that removing neurons rather than connections does not only improve the model size but also the memory footprint and decoding speed. As explained in Sec.~\ref{sec:data-free}, our data-free method is an extension of the approach by Srinivas and Babu~\shortcite{sparsify-datafree}; our extension performs significantly better on NMT networks. Our data-bound method (Sec.~\ref{sec:data-bound}) is inspired by Babaeizadeh et al.~\shortcite{sparsify-noiseout} as we combine neurons with similar activities during training, but we use linear combinations of multiple neurons to compensate for the loss of a neuron rather than merging pairs of neurons.

Using low rank matrices for neural network compression, particularly approximations via SVD, has been studied widely in the literature~\cite{sparsify-lowrank,sparsify-exploitlin,sparsify-svd-replace,sparsify-svd-replace2,sparsify-svd-asr}. These approaches often use low rank matrices to approximate a full rank weight matrix in the original network. In contrast, we shrink an entire linear layer by applying SVD on the product of the incoming and outgoing weight matrices (Sec.~\ref{sec:svd}).

In this paper we mimicked the output of the high performing but cumbersome ensemble by constructing a large unfolded network, and shrank this network afterwards. Another approach, known as {\em knowledge distillation}, uses the large model (the teacher) to generate soft training labels for the smaller student network~\cite{kd-original,kd-hinton}. The student network is trained by minimizing the cross-entropy to the teacher. This idea has been applied to sequence modelling tasks such as machine translation and speech recognition~\cite{kd-asr,kd-nmt,kd-nmt-ensemble}. Our approach can be computationally more efficient as the training set does not have to be decoded by the large teacher network.

Junczys-Dowmunt et al.~\shortcite{averaging2,averaging1} reported gains from averaging the weight matrices of multiple checkpoints of the same training run. However, our attempts to replicate their approach were not successful.   Averaging might work well when the behaviour of corresponding units is similar across networks, but that cannot be guaranteed when networks are trained independently.

\section{Conclusion}

We have described a generic method for improving the decoding speed and BLEU score of single system NMT. Our approach involves unfolding an ensemble of multiple systems into a single large neural network and shrinking this network by removing redundant neurons. Our best results on Japanese-English either yield a gain of 2.2 BLEU compared to the original single NMT network at about the same decoding speed, or a $3.4\times$ CPU decoding speed up with only a minor drop in BLEU.

The current formulation of unfolding works for networks of the same topology as the concatenation of layers is only possible for analogous layers in different networks. Unfolding and shrinking diverse networks could be possible, for example by applying the technique only to the input and output layers or by some other scheme of finding associations between units in different models, but we leave this investigation to future work as models in NMT ensembles in current research usually have the same topology~\cite{sys-wmt16,sys-uedin-wmt16,nmt-char-noseg,sys-neubig-wat16,sys-google,sys-qcri}.

\section*{Acknowledgments}

This work  was  supported  by the U.K.\ Engineering and Physical Sciences Research Council (EPSRC grant EP/L027623/1).

\section*{Appendix: Probabilistic Interpretation of Data-Free and Data-Bound Shrinking}

Data-free and data-bound shrinking can be interpreted as setting the expected difference between network outputs before and after a removal operation to zero under different assumptions.

For simplicity, we focus our probabilistic treatment of shrinking on single layer feedforward networks. Such a network maps an input $\mathbf{x}\in\mathbb{R}^{m_{in}}$ to an output $\mathbf{y}\in\mathbb{R}^{m_{out}}$. The $l$-th output $y_l$ is computed according the following equation

\begin{equation}
\label{eq:appendix-before}
y_l = \sum_{k\in[1,m]} \sigma(\mathbf{x}\mathbf{u}^T_k)V_{k,l}
\end{equation}
where $\mathbf{u}_k\in\mathbb{R}^{m_{in}}$ is the incoming weight vector of the $k$-th hidden neuron (denoted as $U_{:,k}$ in the main paper) and $V\in\mathbb{R}^{m\times m_{out}}$ the outgoing weight matrix of the $m$-dimensional hidden layer. We now remove the $j$-th neuron in the hidden layer and modify the outgoing weights to compensate for the removal:

\begin{equation}
\label{eq:appendix-after}
y_l' = \sum_{k\in[1,m]\setminus \{j\}} \sigma(\mathbf{x}\mathbf{u}^T_k)V'_{k,l}
\end{equation}
where $y_l'$ is the output after the removal operation and $V'\in\mathbb{R}^{m\times m_{out}}$ are the modified outgoing weights. Our goal is to choose $V'$ such that the expected error introduced by removing neuron $j$ is zero:

\begin{equation}
\label{eq:appendix-expect-zero}
\mathbb{E}_\mathbf{x}(y_l-y_l') = 0
\end{equation}

\paragraph{Data-free shrinking}

Data-free shrinking makes two assumptions to satisfy Eq.~\ref{eq:appendix-expect-zero}. First, we assume that the incoming weight vector $\mathbf{u}_j$ can be represented as linear combination of the other weight vectors.

\begin{equation}
\label{eq:appendix-lambda-datafree}
\mathbf{u}_j = \sum_{k\in[1,m]\setminus \{j\}} \lambda_k \mathbf{u}_k
\end{equation}

Second, it assumes that the neuron activation function $\sigma(\cdot)$ is linear. 
Starting with Eqs.~\ref{eq:appendix-before} and~\ref{eq:appendix-after} we can write $\mathbb{E}_{\mathbf{x}}(y_l-y_l')$ as

\begin{equation*}
\begin{array}{@{\hspace{0em}}c@{\hspace{-0.65em}}r@{\hspace{0em}}}
& 
\!\!\!\!\!\!\!\!\!\!\!\!\!\!\!\!\hspace*{1em}\mathbb{E}_\mathbf{x} \Big( \sigma(\mathbf{x}\mathbf{u}_j^T)V_{j,l} 
+ \hspace*{-1em}\underbrace{\sum_{k\in[1,m]\setminus \{j\}} \!\!\!\!\sigma(\mathbf{x}\mathbf{u}_k^T)(V_{k,l} - V'_{k,l})}_{:=R}\Big)\\
\overset{\text{Eq.~\ref{eq:appendix-lambda-datafree}}}{=}& \displaystyle\mathbb{E}_\mathbf{x} \Big( \sigma(\mathbf{x}(\sum_{k\in[1,m]\setminus \{j\}} \lambda_k \mathbf{u}_k)^T)V_{j,l} +R\Big)\\
\overset{\sigma(\cdot)\text{ lin.}}{=}& \displaystyle\mathbb{E}_\mathbf{x} \Big( \sum_{k\in[1,m]\setminus \{j\}} \sigma(\mathbf{x}\mathbf{u}_k^T)\lambda_k V_{j,l}  +R\Big)\\
=& \!\displaystyle\sum_{k\in[1,m]\setminus \{j\}} \!\!\mathbb{E}_\mathbf{x} \Big(\sigma(\mathbf{x}\mathbf{u}_k^T)\Big)(V_{k,l} - V'_{k,l} + \lambda_k V_{j,l}) \\
\end{array}
\end{equation*}

We set this term to zero (and thus satisfy Eq.~\ref{eq:appendix-expect-zero}) by setting each component of the sum to zero.

\begin{equation}
\label{eq:appendix-update}
\forall k\in[1,m]\setminus\{j\}: V_{k,l}' = V_{k,l} + \lambda_k V_{j,l}
\end{equation}
This condition is directly implemented by the update rule in our shrinking algorithm (Eq.~\ref{eq:update-delta} and \ref{eq:update}).

\paragraph{Data-bound shrinking}

Data-bound shrinking does not require linearity in $\sigma(\cdot)$. It rather assumes that the expected value of the neuron activity $j$ is a linear combination of the expected values of the other activities:

\begin{equation}
\label{eq:appendix-lambda-databound}
\mathbb{E}_\mathbf{x}(\sigma(\mathbf{x}\mathbf{u}_j^T)) = \sum_{k\in[1,m]\setminus\{j\}} \lambda_k \mathbb{E}_\mathbf{x}(\sigma(\mathbf{x}\mathbf{u}_k^T))
\end{equation}

$\mathbb{E}_\mathbf{x}(\cdot)$ is estimated using importance sampling:

\begin{equation}
\hat{\mathbb{E}}_\mathbf{x}(\sigma(\mathbf{x}\mathbf{u}_k^T); \mathcal{X})=\frac{1}{|\mathcal{X}|}\sum_{\mathbf{x}'\in\mathcal{X}} \sigma(\mathbf{x}'\mathbf{u}_k^T)
\end{equation}

In practice, the samples in $\mathcal{X}$ are collected in the activity matrix $A$ from Sec.~\ref{sec:data-bound}. We can satisfy Eq.~\ref{eq:appendix-expect-zero} by using the $\lambda$-values from Eq.~\ref{eq:appendix-lambda-databound}, so that $\displaystyle\mathbb{E}_\mathbf{x}(y_l-y_l')$ becomes

\begin{equation*}
\begin{array}{@{\hspace{0em}}c@{\hspace{-0.75em}}r@{\hspace{0em}}}
\overset{\text{Eqs.~\ref{eq:appendix-before},\ref{eq:appendix-after}}}{=}&\hspace{1ex} \displaystyle\mathbb{E}_\mathbf{x} \Big( \sigma(\mathbf{x}\mathbf{u}_j^T)V_{j,l} \\
 &+ \displaystyle\sum_{k\in[1,m]\setminus \{j\}} \sigma(\mathbf{x}\mathbf{u}_k^T)(V_{k,l} - V'_{k,l})\Big)\\
=& \hspace{1em} \displaystyle\mathbb{E}_\mathbf{x} (\sigma(\mathbf{x}\mathbf{u}_j^T)V_{j,l}) \\
&+ \displaystyle\sum_{k\in[1,m]\setminus \{j\}} \mathbb{E}_\mathbf{x}(\sigma(\mathbf{x}\mathbf{u}_k^T))(V_{k,l} - V'_{k,l})\\
\overset{\text{Eq.~\ref{eq:appendix-lambda-databound}}}{=}&\displaystyle\sum_{k\in[1,m]\setminus \{j\}} \!\!\mathbb{E}_\mathbf{x}(\sigma(\mathbf{x}\mathbf{u}_k^T))(V_{k,l} - V'_{k,l}+\lambda_k V_{j,l})\\
\end{array}
\end{equation*}
Again, we set this to zero using Eq.~\ref{eq:appendix-update}.

\bibliography{refs}
\bibliographystyle{emnlp_natbib}

\end{document}